\documentclass[10pt, a4paper]{article}

\usepackage{lrec}
\usepackage{graphicx}
\usepackage{tabularx}
\usepackage{soul}

\usepackage{epstopdf}
\usepackage[latin1,utf8]{inputenc}

\usepackage{hyperref}
\usepackage{xstring}

\usepackage{changepage}
\usepackage{booktabs} %

\usepackage{makecell}
\usepackage{tikz}
\usetikzlibrary{calc}

\title{A Corpus for Detecting High-Context Medical Conditions\\in Intensive Care Patient Notes\\ Focusing on Frequently Readmitted Patients}

\name{Edward T. Moseley\textsuperscript{1,2}, Joy T. Wu\textsuperscript{1,3}, Jonathan Welt\textsuperscript{1,4}, John Foote Jr\textsuperscript{1,5}, \\\large{\textbf{Patrick D. Tyler\textsuperscript{1,6}, David W. Grant\textsuperscript{7}, Eric T. Carlson\textsuperscript{8},}}\\  \large{\textbf{Sebastian Gehrmann\textsuperscript{1,9}, Franck
Dernoncourt\textsuperscript{1,10,11}, Leo Anthony Celi\textsuperscript{1,11}}}}

\address{$^{1}$ MIT Critical Data, Laboratory for Computational Physiology, Cambridge, MA, USA  \\ 
$^{2}$ College of Science and Mathematics, University of Massachusetts, Boston, MA, USA \\
$^{3}$ Harvard T.H. Chan School of Public Health, Cambridge, MA, USA \\ 
$^{4}$ Wellman Center for Photomedicine, Massachusetts General Hospital, Boston, MA, USA \\
$^{5}$ Tufts University School of Medicine, Cambridge, MA, USA \\
$^{6}$ Department of Internal Medicine, Beth Israel Deaconess Medical Center, Boston, MA, USA \\
$^{7}$ Department of Surgery, Division of Plastic and Reconstructive Surgery,\\ Washington University School of Medicine, St. Louis, MO, USA \\
$^{8}$ Philips Research North America, Cambridge, MA, USA\\ 
$^{9}$ Harvard SEAS, Harvard University, Cambridge, MA, USA \\
$^{10}$ Adobe Research, San Jose, CA, USA \\
$^{11}$ Massachusetts Institute of Technology, Cambridge, MA, USA\\
         etmoseley@yahoo.com, joytywu@gmail.com, jonathanwelt@gmail.com, johnfootejr@gmail.com, \\ patrickdtyler@gmail.com, dwgrant101@gmail.com, eric@ds-do.com,  \\ gehrmann@seas.harvard.edu, franck.dernoncourt@adobe.com, lceli@bidmc.harvard.edu}

\abstract{
A crucial step within secondary analysis of electronic health records (EHRs) is to identify the patient cohort under investigation. While EHRs contain medical billing codes that aim to represent the conditions and treatments patients may have, much of the information is only present in the patient notes. Therefore, it is critical to develop robust algorithms to infer patients' conditions and treatments from their written notes.
In this paper, we introduce a dataset for patient phenotyping, a task that is defined as the identification of whether a patient has a given medical condition (also referred to as clinical indication or phenotype) based on their patient note. Nursing Progress Notes and Discharge Summaries from the Intensive Care Unit of a large tertiary care hospital were manually annotated for the presence of several high-context phenotypes relevant to treatment and risk of re-hospitalization. %
This dataset contains 1102 Discharge Summaries and 1000 Nursing Progress Notes. Each Discharge Summary and Progress Note has been annotated by at least two expert human annotators (one clinical researcher and one resident physician). Annotated phenotypes include treatment non-adherence, chronic pain, advanced/metastatic cancer, as well as 10 other phenotypes. This dataset can be utilized for academic and industrial research in medicine and computer science, particularly within the field of medical natural language processing. \\ \newline \Keywords{medical NLP, patient notes, document classification}}

\begin{document}

\maketitleabstract
\begin{tikzpicture}[remember picture, overlay]
\node at ($(current page.north) + (-0in,-0.5in)$) {Accepted as a conference paper at LREC 2020};
\end{tikzpicture}

\section{Introduction and Related Work}

With the widespread adoption of electronic health records (EHRs), medical data are being generated and stored digitally in vast quantities \cite{Henry-16}. While much EHR data are structured and amenable to analysis, there appears to be limited homogeneity in data completeness and quality \cite{pmid22733976}, and it is estimated that the majority of healthcare data are being generated in unstructured, text-based format \cite{pmid23549579}. The generation and storage of these unstructured data come concurrently with policy initiatives that seek to utilize preventative measures to reduce hospital admission and readmission \cite{pmid26910198}.

Chronic illnesses, behavioral factors, and social determinants of health are known to be associated with higher risks of hospital readmission, \cite{pmid30128789} and though behavioral factors and social determinants of health are often determined at the point of care, their identification may not always be curated in structured format within the EHR in the same manner that other factors associated with routine patient history taking and physical examination are \cite{CritData}. Identifying these patient attributes within EHRs in a reliable manner has the potential to reveal actionable associations which otherwise may remain poorly defined.

As EHRs act to streamline the healthcare administration process, much of the data collected and stored in structured format may be those data most relevant to reimbursement and billing, and may not necessarily be those data which were most relevant during the clinical encounter. For example, a diabetic patient who does not adhere to an insulin treatment regimen and who thereafter presents to the hospital with symptoms indicating diabetic ketoacidosis (DKA) will be treated and considered administratively as an individual presenting with DKA, though that medical emergency may have been secondary to non-adherence to the initial treatment regimen in the setting of diabetes. In this instance, any retrospective study analyzing only the structured data from many similarly selected clinical encounters will necessarily then underestimate the effect of treatment non-adherence with respect to hospital admissions.

\begin{table*}[!htb]
\centering

\caption{The thirteen different phenotypes used for our dataset, as well the definition for each phenotype that was used to identify and annotate the phenotype.\vspace{1cm}}

\label{tab:phenotype-description}
\begin{tabularx}{\textwidth}{|p{3.5cm}|X|}
\hline
\textbf{Phenotype} & \textbf{Definition}   \\ \hline


Adv. / Metastatic Cancer &  Cancers with very high or imminent mortality (pancreas, esophagus, stomach, cholangiocarcinoma, brain); mention of distant or multi-organ metastasis, where palliative care would be considered (prognosis $< 6$ months). \newline Example: ``h/o cholangiocarcinoma dx in [DATE] s/p resection, with recent CT showing met cholangiocarcinoma''. \\ \hline

Adv. Heart Disease &  Any consideration for needing a heart transplant; description of severe aortic stenosis (aortic valve area $< 1.0\mathrm{cm}^2$), severe cardiomyopathy, Left Ventricular Ejection Fraction (LVEF) \textless= 30\%. Not sufficient to have a medical history of congestive heart failure (CHF) or myocardial infarction (MI) with stent or coronary artery bypass graft (CABG) as these are too common. \newline Example: ``echo in [DATE] showed EF 30\%''.  \\ \hline

Adv. Lung Disease &  Severe chronic obstructive pulmonary disease (COPD) defined as Gold Stage III-IV, or with a forced expiratory volume during first breath (FEV1) $< 50\%$ of normal, or forced vital capacity (FVC) $< 70\%$, or severe interstitial lung disease (ILD), or Idiopathic pulmonary fibrosis (IPF). \newline  Example: ``Pt has significant obstructive \& restrictive pulmonary disease, on home oxygen''. \\ \hline

Alcohol Abuse & Current/recent alcohol abuse history; still an active problem at time of admission (may or may not be the cause of it). \newline  Example: ``past medical history of alcohol abuse who was diagnosed with alcoholic cirrhosis on this admission''. \\ \hline

Chronic Neurologic \newline Dystrophies &  Any chronic central nervous system (CNS) or spinal cord diseases, included/not limited to: Multiple sclerosis (MS), amyotrophic lateral sclerosis (ALS), myasthenia gravis, Parkinson’s Disease, epilepsy, history of stroke/cerebrovascular accident (CVA) with residual deficits, and various neuromuscular diseases/dystrophies. \newline Example: ``58 yo m w/ multiple sclerosis and seizure disorder''. \\ \hline

Chronic Pain &  Any etiology of chronic pain, including fibromyalgia, requiring long-term opioid/narcotic analgesic medication to control. \newline Example: ``started on a fentanyl patch and continued on prn oxycodone for pain control''. \\ \hline

Dementia & Alzheimer's, alcohol-associated, and other forms of dementia mentioned in the text. \newline  Example: ``pmh sig for Alzheimer's dementia''. \\ \hline 

Depression &  Diagnosis of depression; prescription of anti-depressant medication; or any description of intentional drug overdose, suicide or self-harm attempts. \newline Example: ``Citalopram restarted once taking POs''. \\ \hline 

Developmental Delay &  Includes congenital, genetic and idiopathic disabilities. \newline Example: ``history of Down's syndrome''.\\ \hline

Non Adherence &  Temporary or permanent discontinuation of a treatment, including pharmaceuticals or appointments, without consulting a physician prior to doing so. This includes skipping dialysis appointments or leaving the hospital against medical advice. A patient who sees a physician to discuss adverse events associated with a medication may or may not constitute non-adherence depending on whether or not the treatment was ceased without the physician's consultation. \newline Example: ``in setting of missing dialysis for the past week''.\\ \hline

Obesity & Clinical obesity. BMI $> 30$. Previous history of or being considered for gastric bypass. Insufficient to have abdominal obesity mentioned in physical exam. \newline Example: ``past medical history of morbid obesity with weight greater than 300 pounds''.\\ \hline

Psychiatric disorders &  All psychiatric disorders in DSM-5 classification, including schizophrenia, bipolar and anxiety disorders, other than depression. \newline Example: ``schizophrenia, on effexor and risperdal''.\\ \hline

Substance Abuse &  Include any intravenous drug abuse (IVDU), accidental overdose of psychoactive or narcotic medications,(prescribed or not). Admitting to marijuana use in history is not sufficient. \newline Example: ``toxicology screen was positive for ETOH, benzos, and cocaine''.\\ \hline

\end{tabularx}

\end{table*}


\begin{table*}[!htb]
\centering
\caption{ Data set distribution. The second and fourth columns indicate how many discharge notes and nursing notes contain a given phenotype. The third and fifth columns indicate the corresponding percentages (i.e., the percentage of discharge notes and nursing notes that contain a given phenotype).\vspace{0.5cm}} %
\label{tab:occurrence-counts}
\begin{tabularx}{\textwidth}{|l|c|c|c|X|}
\hline
\textbf{Phenotype} & \textbf{Discharge Notes} & \textbf{\% of Discharges} & \textbf{Nursing Notes} & \textbf{\% of Nursing Notes}  \\ \hline
Adv. / Metastatic Cancer & 74 & 6.72\% &  24  &  2.40\% \\ \hline
Adv. Heart Disease & 172 & 15.61\%  & 35 & 3.50\% \\ \hline
Adv. Lung Disease & 88 &  7.99\%  & 35 & 3.50\% \\ \hline
Alcohol Abuse & 127 & 11.52\%  & 48 & 4.80\% \\ \hline
Chronic Neurologic Dystrophies & 187 & 16.97\%  & 53 & 5.30\% \\ \hline
Chronic Pain & 150 & 13.61\%  & 39 & 3.90\% \\ \hline
Dementia & 43 & 3.90\%  & 22 & 2.20\% \\ \hline 
Depression & 267 & 24.23\%  & 54 & 5.40\% \\ \hline 
Developmental Delay  & 14 & 1.27\%  & 5 & 0.50\% \\ \hline 
Non Adherence & 78 & 7.08\% &  35 & 3.50\%  \\ \hline 
Obesity & 72 & 6.53\% & 12  & 1.20\%   \\ \hline
Psychiatric disorders & 167 & 15.15\% & 39  & 3.90\%  \\ \hline
Substance Abuse & 92 & 8.35\% & 20  & 2.00\%  \\ \hline
\end{tabularx}
\end{table*}

\begin{table*}[!htb]
\centering
\caption{Patient note statistics. IQR stands for interquartile range.\vspace{0.5cm}}
\label{tab:token-counts}
\begin{tabularx}{\textwidth}{|X|c|c|}
\hline
 & \textbf{Discharge Notes} & \textbf{Nursing Notes}  \\ \hline
Number of patient notes  & 1102 & 1000  \\ \hline
Character count, median [IQR] & 10156.00 [7443.00, 13830.75] & 1232.00 [723.00, 1810.25]
\\ \hline
Token count, median [IQR] & 1417.50 [1046.75, 1926.00] & 208.00 [120.00, 312.00]  \\ \hline
Character count (punctuation \& digits removed), median [IQR] & 8223.00 [6045.25, 11216.25] &   1053.00 [609.75, 1534.25]\\ \hline
Token count (punctuation \& digits removed), median [IQR]
 & 1336.50 [971.25, 1817.50] &  191.50 [111.00, 282.00] \\ \hline

\end{tabularx}
\end{table*}

While this form of high context information may not be found in the structured EHR data, it may be accessible in patient notes, including nursing progress notes and discharge summaries, particularly through the utilization of natural language processing (NLP) technologies. \cite{pmid30427267}, \cite{pmid31160009} Given progress in NLP methods, we sought to address the issue of unstructured clinical text by defining and annotating clinical phenotypes in text which may otherwise be prohibitively difficult to discern in the structured data associated with the text entry. For this task, we chose the notes present in the publicly available MIMIC database~\cite{pmid27219127}.

Given the MIMIC database as substrate and the aforementioned policy initiatives to reduce unnecessary hospital readmissions, as well as the goal of providing structure to text, we elected to focus on patients who were frequently readmitted to the ICU \cite{pmid27441405}. In particular, a patient who is admitted to the ICU more than three times in a single year. By defining our cohort in this way we sought to ensure we were able to capture those characteristics unique to the cohort in a manner which may yield actionable intelligence on interventions to assist this patient population.

\section{Data Characteristics}

We have created a dataset of discharge summaries and nursing notes, all in the English language, with a focus on frequently readmitted patients, labeled with 15 clinical patient phenotypes believed to be associated with risk of recurrent Intensive Care Unit (ICU) readmission per our domain experts (co-authors LAC, PAT, DAG) as well as the literature.~\cite{pmid22009101} \cite{pmid22086871} \cite{pmid24515422} 

	Each entry in this database of consists of a Subject Identifier (integer), a Hospital Admission Identifier (integer), Category (string), Text (string), 15 Phenotypes (binary) including ``None'' and ``Unsure'', Batch Date (string), and Operators (string). These variables are sufficient to use the data set alone, or to join it to the MIMIC-III database by Subject Identifier or Hospital Admission Identifier for additional patient-level or admission-level data, respectively.	The MIMIC database~\cite{pmid27219127} was utilized to extract Subject Identifiers, Hospital Admission Identifiers, and Note Text.
	
	Annotated discharge summaries had a median token count of 1417.50 (Q1-Q3: 1046.75 - 1926.00) with a vocabulary of 26454 unique tokens, while nursing notes had a median count of 208 (Q1-Q3: 120 - 312) with a vocabulary of 12865 unique tokens.

\begin{figure*}[!htp]
    \centering
    \includegraphics[width=18cm]{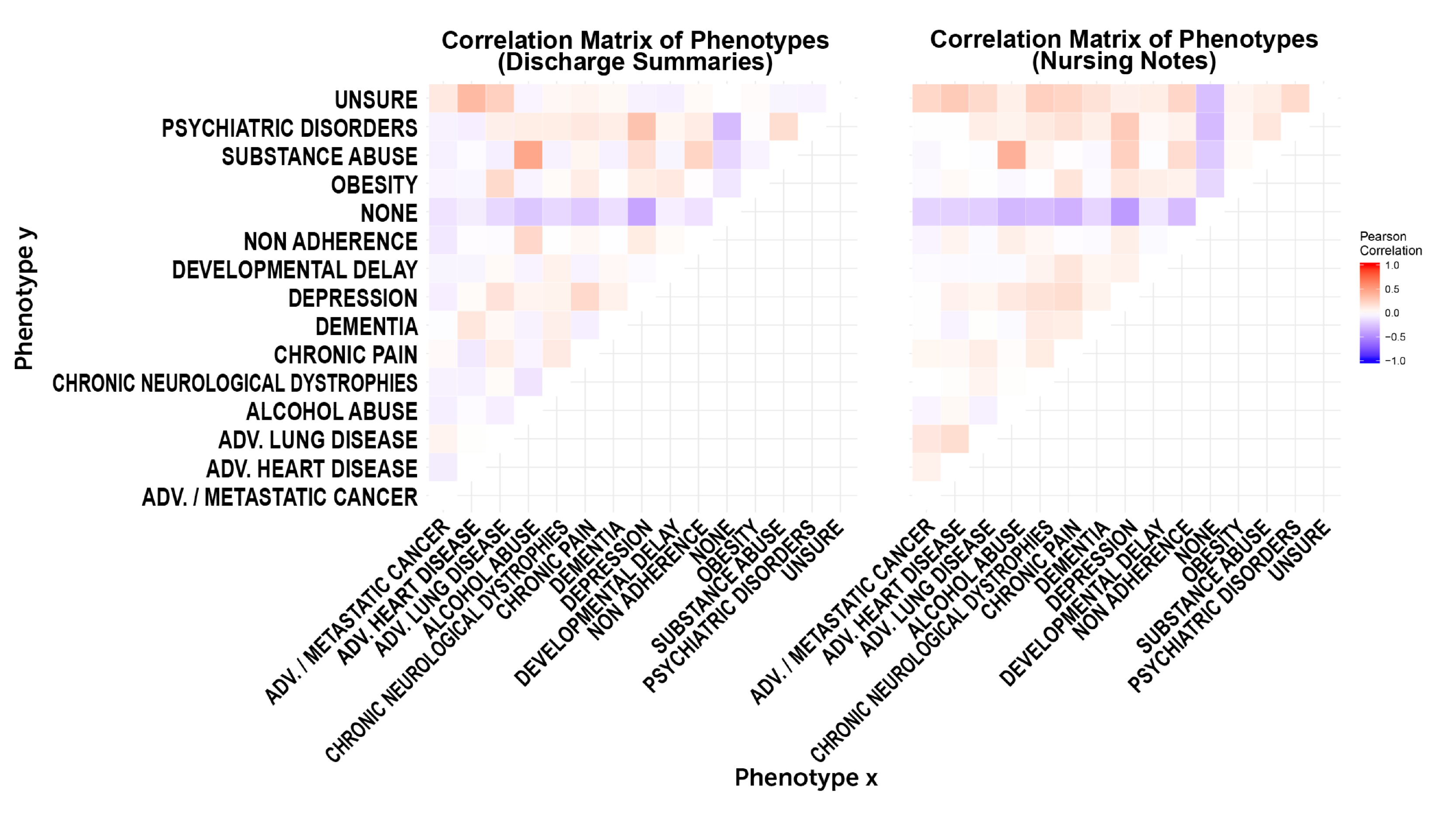}
    \caption{\vspace{0.5cm}Correlation matrices of phenotypes for Nursing Notes and Discharge Summaries.}
    \label{fig:corrmat}
\end{figure*}

	Table~\ref{tab:phenotype-description} defines each of the considered clinical patient phenotypes. Table~\ref{tab:occurrence-counts} counts the occurrences of these phenotypes across patient notes and Figure~\ref{fig:corrmat} contains the corresponding correlation matrix. Lastly, Table~\ref{tab:token-counts} presents an overview of some descriptive statistics on the patient notes' lengths.

\section{Methods}

Clinical researchers teamed with junior medical residents in collaboration with more senior intensive care physicians to carry out text annotation over the period of one year~\cite{wu2018behind}. %
Operators were grouped to facilitate the annotation of notes in duplicate, allowing for cases of disagreement between operators. The operators within each team were instructed to work independently on note annotation. Clinical texts were annotated in batches which were time-stamped on their day of creation, when both operators in a team completed annotation of a batch, a new batch was created and transferred to them.

Two groups (group 1: co-authors ETM \& JTW; group 2: co-authors JW \& JF) of two operator pairs of one clinical researcher and one resident physician (who had previously taken the MCAT\textsuperscript{\textregistered}) first annotated nursing notes and then discharge summaries. Everyone was first trained on the high-context phenotypes to look for as well as their definitions by going through a number of notes in a group. A total of 13 phenotypes were considered for annotation, and the label ``unsure'' was used to indicate that an operator would like to seek assistance determining the presence of an phenotype from a more senior physician. Annotations for phenotypes required explicit text in the note indicating the phenotype, but as a result of the complexity of certain phenotypes there was no specific dictionary of terms, or order in which the terms appeared, required for a phenotype to be considered present.

\section{Limitations}

There exist a few limitations to this database. These data are unique to Beth Israel Deaconess Medical Center (BIDMC), and models resulting from these data may not generalize to notes generated at other hospitals. Admissions to hospitals not associated with BIDMC will not have been captured, and generalizability is limited due to the limited geographic distribution of patients which present to the hospital.

We welcome opportunities to continue to expand this dataset with additional phenotypes sought in the unstructured text, patient subsets, and text originating from different sources, with the goal of expanding the utility of NLP methods to further structure patient note text for retrospective analyses.

\section{Technical Validation}

All statistics and tabulations were generated and performed with R Statistical Software version 3.5.2. \cite{R_lang} Cohen's Kappa~\cite{cohen1960coefficient} was calculated for each phenotype and pair of annotators for which precisely two note annotations were recorded. Table~\ref{tab:kappas} summarizes the calculated Cohen's Kappa coefficients.

\begin{table}[!htb]
\centering
\caption{Tabulated Cohen's Kappa values for each on the two annotator pairs: the first pair is ETM \& JTW, and the second pair is JF \& JW.}
\label{tab:kappas}
\begin{tabularx}{8.5cm}{|X|c|c|}
\hline
  \textbf{Phenotype} & \textbf{ETM \& JTW} & \textbf{JF \& JW}  \\ \hline
Adv. / Metastatic Cancer & 0.834 & 0.869  \\ \hline
Adv. Heart Disease  & 0.820 &  0.561 \\ \hline
Adv. Lung Disease  & 0.805 &  0.654 \\ \hline
Alcohol Abuse  & 0.856 & 0.842  \\ \hline
Chronic Neurologic Dystrophies  & 0.714 &  0.700  \\ \hline
Chronic Pain & 0.833 & 0.731  \\ \hline
Dementia  & 0.952 & 0.947  \\ \hline
Depression  & 0.853 & 0.945  \\ \hline
Developmental Delay  & 1.000 & 0.869  \\ \hline
Non Adherence  & 0.778 & 0.694  \\ \hline
Obesity  & 0.941 &  0.968 \\ \hline
Psychiatric disorders  & 0.908 &  0.940 \\ \hline
Substance Abuse & 0.862 &  0.882 \\ \hline
\end{tabularx}
\end{table}

\section{Usage Notes}

	As this corpus of annotated patient notes comprises original healthcare data which contains protected health information (PHI) per The Health Information Portability and Accountability Act of 1996 (HIPAA)~\cite{act1996health} and can be joined to the MIMIC-III database, individuals who wish to access to the data must satisfy all requirements to access the data contained within MIMIC-III. To satisfy these conditions, an individual who wishes to access the database must take a ``Data or Specimens Management'' course, as well as sign a user agreement, as outlined on the MIMIC-III database webpage, where the latest version of this database will be hosted as ``Annotated Clinical Texts from MIMIC''\footnote{\url{https://mimic.physionet.org}}~\cite{ned2020data}. This corpus can also be accessed on GitHub\footnote{\url{https://github.com/EdwardMoseley/ACTdb}} after completing all of the above requirements. %

\section{Baselines}

In  the section, we present the performance of two well-established baseline models to automatically infer the phenotype based on the patient note, which we approach as a multi-label, multi-class text classification task~\cite{gehrmann2018comparing}. Each of the baseline model is a binary classifier indicating whether a given phenotype is present in the input patient note. As a result, we train a separate model for each phenotype.

\paragraph{Bag of Words + Logistic Regression} We convert each patient note into a bag of words, and give as input to a logistic regression.

\paragraph{Convolutional Neural Network (CNN)} We follow the CNN architecture proposed by \newcite{collobert2011natural} and \newcite{kim2014convolutional}. We use the convolution widths from $1$ to $4$, and for each convolution width we set the number of filters to $100$. We use dropout with a probability of $0.5$ to reduce overfitting~\cite{srivastava2014dropout}. The trainable parameters were initialized using a uniform distribution from $-0.05$ to $0.05$. The model was optimized with adadelta~\cite{zeiler2012adadelta}. We use word2vec~\cite{mikolov2013distributed} as the word embeddings, which we pretrain on all the notes of MIMIC III v3. 

Table~\ref{tab:results} presents the performance of the two baseline models (F1-score).

\begin{table}[!htb]
\centering
\caption{Results of the two baseline models (F1-score in percentage). BoW stands for Bag of Words.}
\label{tab:results}
\begin{tabularx}{8.5cm}{|X|c|c|}
\hline
  \textbf{Phenotype} & \textbf{BoW} & \textbf{CNN}  \\ \hline
Adv. / Metastatic Cancer & 56 &  74 \\ \hline
Adv. Heart Disease  & 44 &  75 \\ \hline
Adv. Lung Disease  & 24 &  48 \\ \hline
Alcohol Abuse  & 67 &  76 \\ \hline
Chronic Neurologic Dystrophies  & 46 &  69 \\ \hline
Chronic Pain  & 41 & 49  \\ \hline
Depression  & 58 & 84  \\ \hline
Obesity  & 30 &  95 \\ \hline
Psychiatric disorders  & 50 &  83 \\ \hline
Substance Abuse & 56 &  75 \\ \hline
\end{tabularx}
\end{table}

\section{Conclusion}

In this paper we have presented a new dataset containing discharge summaries and nursing progress notes, focusing on frequently readmitted patients and high-context social determinants of health, and originating from a large tertiary care hospital. Each patient note was annotated by at least one clinical researcher and one resident physician for 13 high-context patient phenotypes.

Phenotype definitions, dataset distribution, patient note statistics, inter-operator error, and the results of baseline models were reported to demonstrate that the dataset is well-suited for the development of both rule-based and statistical models for patient phenotyping. %
We hope that the release of this dataset will accelerate the development of algorithms for patient phenotyping, which in turn would significantly help medical research progress faster.

\section{Acknowledgements}

The authors would like to acknowledge Kai-ou Tang and William Labadie-Moseley for assistance in the development of a graphical user interface for text annotation. We would also like to thank Philips Healthcare, The Laboratory of Computational Physiology at The Massachusetts Institute of Technology, and staff at the Beth Israel Deaconess Medical Center, Boston, for supporting the MIMIC-III database, from which these data were derived.

\section{Bibliographical References}
\label{main:ref}

\bibliographystyle{lrec}
\bibliography{lrec2020W-xample}

\begin{thebibliography}{}

\bibitem[\protect\citename{Act}1996]{act1996health}
Act, A.
\newblock (1996).
\newblock Health insurance portability and accountability act of 1996.
\newblock {\em Public law}, 104:191.

\bibitem[\protect\citename{Chan \bgroup et al.\egroup }2019]{pmid30427267}
Chan, A., Chien, I., Moseley, E., Salman, S., Kaminer~Bourland, S., Lamas, D.,
  Walling, A.~M., Tulsky, J.~A., and Lindvall, C.
\newblock (2019).
\newblock {{D}eep learning algorithms to identify documentation of serious
  illness conversations during intensive care unit admissions}.
\newblock {\em Palliat Med}, 33(2):187--196, 02.

\bibitem[\protect\citename{Cohen}1960]{cohen1960coefficient}
Cohen, J.
\newblock (1960).
\newblock A coefficient of agreement for nominal scales.
\newblock {\em Educational and psychological measurement}, 20(1):37--46.

\bibitem[\protect\citename{Collobert \bgroup et al.\egroup
  }2011]{collobert2011natural}
Collobert, R., Weston, J., Bottou, L., Karlen, M., Kavukcuoglu, K., and Kuksa,
  P.
\newblock (2011).
\newblock Natural language processing (almost) from scratch.
\newblock {\em Journal of Machine Learning Research}, 12(Aug):2493--2537.

\bibitem[\protect\citename{Gehrmann \bgroup et al.\egroup
  }2018]{gehrmann2018comparing}
Gehrmann, S., Dernoncourt, F., Li, Y., Carlson, E.~T., Wu, J.~T., Welt, J.,
  Foote~Jr, J., Moseley, E.~T., Grant, D.~W., Tyler, P.~D., et~al.
\newblock (2018).
\newblock Comparing deep learning and concept extraction based methods for
  patient phenotyping from clinical narratives.
\newblock {\em PloS one}, 13(2).

\bibitem[\protect\citename{J.~Henry and Patel}2016]{Henry-16}
J.~Henry, Y.~Pylypchuk, T.~S. and Patel, V.
\newblock (2016).
\newblock Adoption of electronic health record systems among u.s. non-federal
  acute care hospitals: 2008-2015.
\newblock Technical report, Office of the National Coordinator for Health
  Information Technology, Washington, DC, May.

\bibitem[\protect\citename{Johnson \bgroup et al.\egroup }2016]{pmid27219127}
Johnson, A.~E., Pollard, T.~J., Shen, L., Lehman, L.~W., Feng, M., Ghassemi,
  M., Moody, B., Szolovits, P., Celi, L.~A., and Mark, R.~G.
\newblock (2016).
\newblock {{M}{I}{M}{I}{C}-{I}{I}{I}, a freely accessible critical care
  database}.
\newblock {\em Sci Data}, 3:160035, May.

\bibitem[\protect\citename{Kangovi \bgroup et al.\egroup }2014]{pmid24515422}
Kangovi, S., Mitra, N., Grande, D., White, M.~L., McCollum, S., Sellman, J.,
  Shannon, R.~P., and Long, J.~A.
\newblock (2014).
\newblock {{P}atient-centered community health worker intervention to improve
  posthospital outcomes: a randomized clinical trial}.
\newblock {\em JAMA Intern Med}, 174(4):535--543, Apr.

\bibitem[\protect\citename{Kansagara \bgroup et al.\egroup }2011]{pmid22009101}
Kansagara, D., Englander, H., Salanitro, A., Kagen, D., Theobald, C., Freeman,
  M., and Kripalani, S.
\newblock (2011).
\newblock {{R}isk prediction models for hospital readmission: a systematic
  review}.
\newblock {\em JAMA}, 306(15):1688--1698, Oct.

\bibitem[\protect\citename{Kansagara \bgroup et al.\egroup }2012]{pmid22086871}
Kansagara, D., Ramsay, R.~S., Labby, D., and Saha, S.
\newblock (2012).
\newblock {{P}ost-discharge intervention in vulnerable, chronically ill
  patients}.
\newblock {\em J Hosp Med}, 7(2):124--130, Feb.

\bibitem[\protect\citename{Kim}2014]{kim2014convolutional}
Kim, Y.
\newblock (2014).
\newblock Convolutional neural networks for sentence classification.
\newblock {\em arXiv preprint arXiv:1408.5882}.

\bibitem[\protect\citename{Lai \bgroup et al.\egroup }2016]{CritData}
Lai, Y., Moseley, E., and Salgueiro, F., (2016).
\newblock {\em MIT Critical Data, Secondary Analysis of Electronic Health
  Records}, chapter {{I}ntegrating {N}on-clinical {D}ata with {E}{H}{R}s},
  pages 51--60.
\newblock Springer International Publishing.

\bibitem[\protect\citename{Mikolov \bgroup et al.\egroup
  }2013]{mikolov2013distributed}
Mikolov, T., Sutskever, I., Chen, K., Corrado, G.~S., and Dean, J.
\newblock (2013).
\newblock Distributed representations of words and phrases and their
  compositionality.
\newblock In {\em Advances in neural information processing systems}, pages
  3111--3119.

\bibitem[\protect\citename{Moon \bgroup et al.\egroup }2019]{pmid31160009}
Moon, S., Liu, S., Scott, C.~G., Samudrala, S., Abidian, M.~M., Geske, J.~B.,
  Noseworthy, P.~A., Shellum, J.~L., Chaudhry, R., Ommen, S.~R., Nishimura,
  R.~A., Liu, H., and Arruda-Olson, A.~M.
\newblock (2019).
\newblock {{A}utomated extraction of sudden cardiac death risk factors in
  hypertrophic cardiomyopathy patients by natural language processing}.
\newblock {\em Int J Med Inform}, 128:32--38, 08.

\bibitem[\protect\citename{Moseley \bgroup et al.\egroup }2020]{ned2020data}
Moseley, E.~T., Celi, L.~A., Wu, J.~T., and Dernoncourt, F.
\newblock (2020).
\newblock Phenotype annotations for patient notes in the {MIMIC-III} database.
\newblock {\em PhysioNet}.

\bibitem[\protect\citename{Murdoch and Detsky}2013]{pmid23549579}
Murdoch, T.~B. and Detsky, A.~S.
\newblock (2013).
\newblock {{T}he inevitable application of big data to health care}.
\newblock {\em JAMA}, 309(13):1351--1352, Apr.

\bibitem[\protect\citename{{R Core Team}}2018]{R_lang}
{R Core Team}, (2018).
\newblock {\em R: A Language and Environment for Statistical Computing}.
\newblock R Foundation for Statistical Computing, Vienna, Austria.

\bibitem[\protect\citename{Ryan \bgroup et al.\egroup }2015]{pmid27441405}
Ryan, J., Hendler, J., and Bennett, K.~P.
\newblock (2015).
\newblock {{U}nderstanding {E}mergency {D}epartment 72-{H}our {R}evisits
  {A}mong {M}edicaid {P}atients {U}sing {E}lectronic {H}ealthcare {R}ecords}.
\newblock {\em Big Data}, 3(4):238--248, Dec.

\bibitem[\protect\citename{Srivastava \bgroup et al.\egroup
  }2014]{srivastava2014dropout}
Srivastava, N., Hinton, G., Krizhevsky, A., Sutskever, I., and Salakhutdinov,
  R.
\newblock (2014).
\newblock Dropout: A simple way to prevent neural networks from overfitting.
\newblock {\em The Journal of Machine Learning Research}, 15(1):1929--1958.

\bibitem[\protect\citename{Virapongse and Misky}2018]{pmid30128789}
Virapongse, A. and Misky, G.~J.
\newblock (2018).
\newblock {{S}elf-{I}dentified {S}ocial {D}eterminants of {H}ealth during
  {T}ransitions of {C}are in the {M}edically {U}nderserved: a {N}arrative
  {R}eview}.
\newblock {\em J Gen Intern Med}, 33(11):1959--1967, 11.

\bibitem[\protect\citename{Weiskopf and Weng}2013]{pmid22733976}
Weiskopf, N.~G. and Weng, C.
\newblock (2013).
\newblock {{M}ethods and dimensions of electronic health record data quality
  assessment: enabling reuse for clinical research}.
\newblock {\em J Am Med Inform Assoc}, 20(1):144--151, Jan.

\bibitem[\protect\citename{Wu \bgroup et al.\egroup }2018]{wu2018behind}
Wu, J.~T., Dernoncourt, F., Gehrmann, S., Tyler, P.~D., Moseley, E.~T.,
  Carlson, E.~T., Grant, D.~W., Li, Y., Welt, J., and Celi, L.~A.
\newblock (2018).
\newblock Behind the scenes: A medical natural language processing project.
\newblock {\em International journal of medical informatics}, 112:68--73.

\bibitem[\protect\citename{Zeiler}2012]{zeiler2012adadelta}
Zeiler, M.~D.
\newblock (2012).
\newblock Adadelta: an adaptive learning rate method.
\newblock {\em arXiv preprint arXiv:1212.5701}.

\bibitem[\protect\citename{Zuckerman \bgroup et al.\egroup }2016]{pmid26910198}
Zuckerman, R.~B., Sheingold, S.~H., Orav, E.~J., Ruhter, J., and Epstein, A.~M.
\newblock (2016).
\newblock {{R}eadmissions, {O}bservation, and the {H}ospital {R}eadmissions
  {R}eduction {P}rogram}.
\newblock {\em N. Engl. J. Med.}, 374(16):1543--1551, Apr.

\end{thebibliography}

\end{document}